\pdfoutput=1
\documentclass[11pt]{article}

\usepackage[final]{acl}

\usepackage{times}
\usepackage{latexsym}
\usepackage[inkscapearea=page]{svg}
\usepackage{tabularx}
\usepackage{hyperref}
\usepackage{multirow}
\usepackage{bm}
\usepackage{soul}
\usepackage{makecell}
\usepackage{booktabs}
\usepackage{float}
\usepackage{enumitem}

\usepackage[T1]{fontenc}
\usepackage[utf8]{inputenc}

\usepackage{microtype}

\usepackage{inconsolata}

\usepackage{graphicx}

\title{Enhancing Pre-Trained Generative Language Models with Question Attended Span Extraction on Machine Reading Comprehension}
\author{
    Lin Ai\quad Zheng Hui\quad Zizhou Liu\quad Julia Hirschberg  \\
    Columbia University, New York, NY \\
    \texttt{\{lin.ai, julia\}@cs.columbia.edu} \\
    \texttt{\{zh2483, zl2889\}@columbia.edu}
}

\begin{document}
\maketitle

\begin{abstract}

% To address the challenges of out-of-control generation in generative models for machine reading comprehension (MRC), we introduce the \textbf{Q}uestion-\textbf{A}ttended \textbf{S}pan \textbf{E}xtraction (\textit{QASE}) module. Integrated during the fine-tuning of pre-trained generative language models (PLMs), \textit{QASE} enables these PLMs to match SOTA extractive methods and outperform leading LLMs like GPT-4 in MRC tasks, without significant increases in computational costs. \footnote{Our code is available at \href{https://anonymous.4open.science/r/QASE-7753/README.md}{this anonymous repo link}.}

Machine Reading Comprehension (MRC) poses a significant challenge in the field of Natural Language Processing (NLP). While mainstream MRC methods predominantly leverage extractive strategies using encoder-only models such as BERT, generative approaches face the issue of \textit{out-of-control generation} -- a critical problem where answers generated are often incorrect, irrelevant, or unfaithful to the source text. To address these limitations in generative models for extractive MRC, we introduce the \textbf{Q}uestion-\textbf{A}ttended \textbf{S}pan \textbf{E}xtraction (\textit{QASE}) module. Integrated during the fine-tuning phase of pre-trained generative language models (PLMs), \textit{QASE} significantly enhances their performance, allowing them to surpass the extractive capabilities of advanced Large Language Models (LLMs) such as GPT-4 in few-shot settings. Notably, these gains in performance do not come with an increase in computational demands. The efficacy of the \textit{QASE} module has been rigorously tested across various datasets, consistently achieving or even surpassing state-of-the-art (SOTA) results, thereby bridging the gap between generative and extractive models in extractive MRC tasks. Our code is available at \href{https://github.com/lynneeai/QASE.git}{this GitHub repository.}

\end{abstract}
\section{Introduction}

Extractive Machine Reading Comprehension (MRC), also referred to as text-grounded question answering (QA) \cite{wang2022survey}, involves presenting a model with a text passage and a question, requiring it to formulate an answer based solely on the given text. This can be achieved either by identifying a specific span within the text or by generating a concise answer. Extractive MRC poses a significant challenge within the domain of Natural Language Processing (NLP). Predominant strategies for addressing extractive MRC employ extractive methods, which typically extract pertinent text snippets from a broader context in response to a query \cite{wang-etal-2018-multi-granularity, yan2019deep, chen2020question}. However, the most precise answers in practical settings often span multiple text passages or necessitate inferential reasoning that extends beyond the surface-level content \cite{li-etal-2021-addressing-semantic}. Therefore, there is a compelling necessity to integrate generative models alongside extractive approaches to enhance the robustness, versatility, and comprehensiveness of solutions in this field.

\begin{figure}
    \centering
    \includegraphics[width=0.48\textwidth]{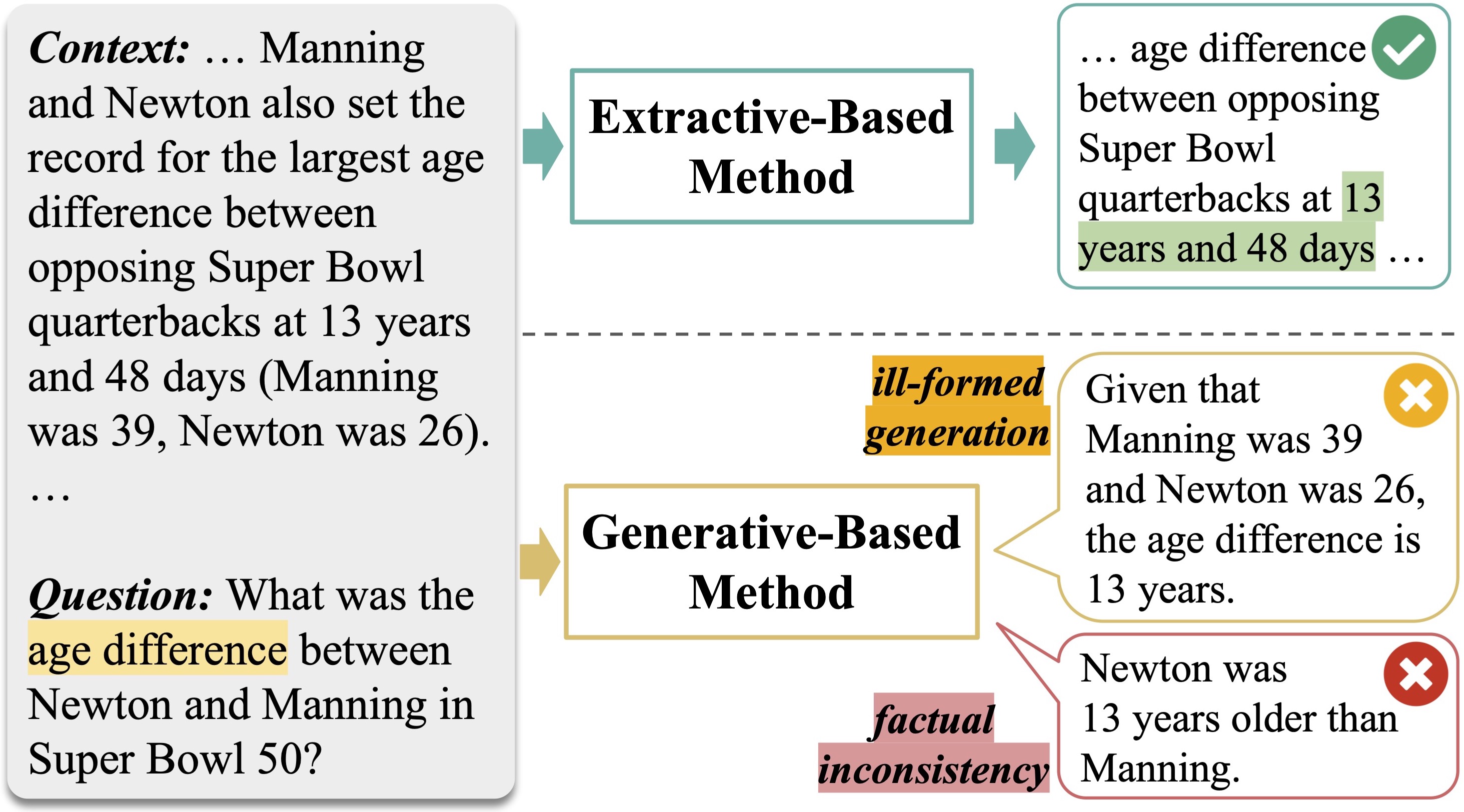}
    \caption{\textit{Out-of-control generation} issue in generative-based methods.}
    \label{fig:extractive_vs_generative}
\end{figure}

Yet, generative models often fall short in extractive MRC tasks due to a phenomenon known as \textit{out-of-control generation} \cite{li-etal-2021-addressing-semantic}, which encompasses two primary issues, as illustrated in Figure \ref{fig:extractive_vs_generative}: \textbf{(a)} ill-formed generations that include incomplete or redundant phrases, and \textbf{(b)} factual inconsistencies that diverge from the intended information. Our research aim to bridge the performance gap between generative and extractive models in extractive MRC tasks by tackling the \textit{out-of-control generation} issue. We introduce the lightweight \textbf{Q}uestion-\textbf{A}ttended \textbf{S}pan \textbf{E}xtraction (\textit{QASE}) module. This module is integrated during the fine-tuning of various open-source generative pre-trained language models (PLMs) across multiple MRC datasets to enhance the reliability and accuracy of the generated answers.

Our key contributions are outlined as follows:

\begin{itemize}
    
\item[1.] We develop the \textit{QASE} module to enhance the quality and factual accuracy of answers generated by fine-tuned generative PLMs, achieving performance on par with state-of-the-art (SOTA) extractive methods and surpassing that of advanced Large Language Models (LLMs) such as GPT-4 in few-shot settings.

\item[2.] \textit{QASE} enhances model performance without imposing significant additional computational demands, offering a cost-effective solution.

\end{itemize}

% This novel approach not only addresses critical shortcomings in generative MRC methodologies but also sets a new benchmark in the efficiency and efficacy of machine comprehension models.
\section{Related Work}
\label{sec:related_work}

% Most \textbf{current studies on MRC} involve predicting the start and end positions of the answer spans from a given context \cite{ohsugi-etal-2019-simple, lan2019albert, bachina-etal-2021-ensemble, chen2022good} using encoder-only PLM models such as BERT and XLM-Roberta. To handle the multi-span setting, some studies frame the problem as a sequence tagging task \cite{segal-etal-2020-simple}, and others explore ways to combine models with different tasks \cite{hu-etal-2019-multi, lee2023liquid, zhang2023many}. While these extractive-based methods mainly utilize encoder-only models, there is also research focuses on using generative language models \cite{yang2020multi, li-etal-2021-addressing-semantic, su-etal-2022-read}. 

\paragraph{Extractive MRC} 
Recent MRC research predominantly focuses on extractive question answering using encoder-only PLMs like BERT and XLM-Roberta, predicting the start and end positions of answers directly from the context \cite{ohsugi-etal-2019-simple, lan2019albert, bachina-etal-2021-ensemble, chen2022good}. For multi-span answers, \citet{segal-etal-2020-simple} treat this as a sequence tagging task, while others \cite{hu-etal-2019-multi, lee2023liquid, zhang2023many} use hybrid approaches to enhance performance on complex MRC problems. Beyond extractive methods, there is growing interest in applying generative language models for extractive MRC \cite{yang2020multi, li-etal-2021-addressing-semantic, jiang2022understanding, su-etal-2022-read}, which generate answers by reformulating information across the context. \citet{xu2021attention} adopt a similar approach to ours by adding a span extraction auxiliary task to guide text generation. However, this method does not focus sufficiently on the queried questions, which may reduce the accuracy of span extraction.

\paragraph{Retrieval-augmented text generation (RAG)} 
RAG augments the input of PLMs with in-domain \cite{gu2018search, weston-etal-2018-retrieve, saha-srihari-2023-argu} or external knowledge \cite{su2021prototype, xiao-etal-2021-transductive} to control the quality and factual consistency of generated content. It has become a new text generation paradigm in many NLP tasks \cite{li2022survey, ai2024defending}, such as dialogue response generation \cite{wu2021controllable, liu-etal-2023-recap} and machine translation \cite{he-etal-2021-fast, zhu-etal-2023-ink}. However, RAG is typically utilized in scenarios where document retrieval is necessary to reduce input context window \cite{Chen_Lin_Han_Sun_2024,ram_in_context_rag_2023}, whereas selective MRC often requires accessing information beyond the immediate context. Our approach diverges from RAG as it directly fine-tunes the weights of the PLMs rather than altering the input to the PLMs with additional information.

\paragraph{Controllable Text Generation} 
Significant progress has been made in controllable text generation. \citet{gururangan-etal-2020-dont} fine-tune language models on domain-adaptive text to customize generated content attributes. Other methods include reinforcement learning \cite{li2024reinforcement}, contrastive learning \cite{zheng2023click}, and control codes for fine-tuning PLMs \cite{keskar2019ctrl}. Some approaches modify the probability distribution of PLMs, such as \citet{liu-etal-2021-dexperts} using two smaller “expert” models, and \citet{yang-klein-2021-fudge} conditioning generation with a “future discriminator.” \citet{huang-etal-2023-extensible} explore multi-aspect text generation with trainable gates for enhanced control. Our proposed module, \textit{QASE}, represents a novel adaptation of controlled text generation tailored to the specific challenges of MRC, with a focus on the precision and relevance of generated answers. Unlike methods that modify the overall generative process through complex architectural alterations or additional learning mechanisms, \textit{QASE} directly utilizes the question and context to guide inferences.
\section{Method}
This section presents our proposed \textit{QASE} module and the multi-task fine-tuning strategy we employ.

\vspace{-0.2cm}
\begin{figure*}[h]
    \centering
    \includegraphics[width=0.95\textwidth]{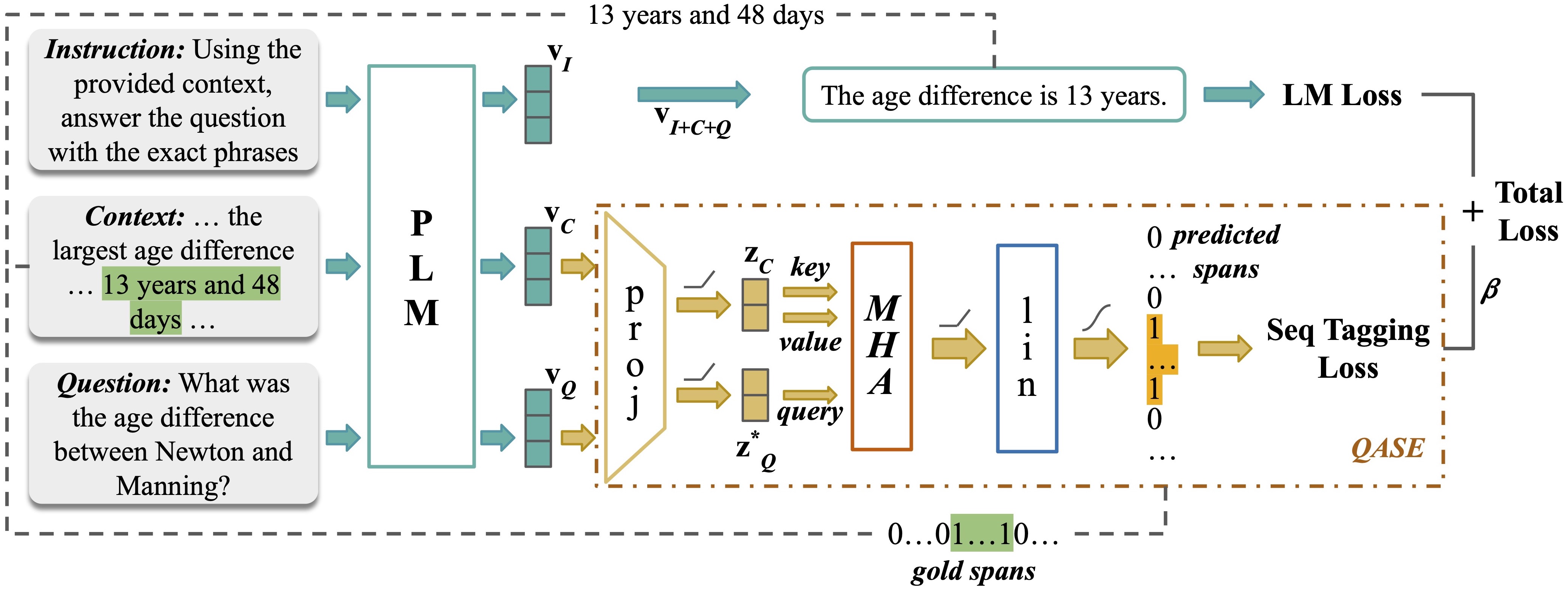}
    \caption{Architecture of the \textit{QASE}-enhanced model. Here, $z^*_Q$ represents the averaged embedding of question tokens, expanded to match the length of $z_C$.}
    \label{fig:model}
    % \vspace{-0.2cm}
\end{figure*}

\subsection{Question-Attended Span Extraction}
\label{subsec:span_extraction}

To guide text generation, we employ the \textit{QASE} module, a question-attended span extraction tool, during the fine-tuning of generative PLMs. \textit{QASE} directs model focus to potential answer spans within the original text. We frame span extraction as a sequence tagging task using the Inside-Outside (IO) tagging schema. In this schema, each token is labeled as `inside' (\textbf{\textit{I}}) if it falls within a relevant span, or `outside' (\textbf{\textit{O}}) otherwise. This approach effectively handles both single- and multi-span extractions and has shown to perform on par with or better than the well-known BIO format \cite{huang2015bidirectional}, as demonstrated by \citet{segal-etal-2020-simple}.

The model architecture is depicted in Figure \ref{fig:model}. Initially, a context and question pair along with an instruction are tokenized and input into the PLM. The resultant hidden states from the PLM are then transformed through projection layers to generate embeddings $z_i = ReLU(W_{proj}v_i + b_{proj})$, where $v_i \in R^d$ represents the hidden state of the $i^{th}$ token from the PLM output.

To capture the relationship of context tokens to specific questions, we utilize a multi-head attention mechanism (\textbf{\textit{MHA}}). Each attention head targets different aspects of the context in relation to the question, treating question embeddings as queries and context embeddings as keys and values. Specifically, for each question-context pair, we compute a mean question embedding by averaging the embeddings of question tokens, which is then expanded to align with the length of the context sequence. This expanded question embedding, $z^*_Q$, serves as the query in the \textbf{\textit{MHA}}, with the context embedding, $z_C$, acting as both the key and the value. This mechanism allows the derived representation of each token in the context to encapsulate its relevance in relation to the posed question.

In conclusion, the \textit{QASE} module processes the projected embeddings $z_C$ and $z^*_Q$ through the \textbf{\textit{MHA}} mechanism, followed by a linear and a softmax layer to calculate the probability that each context token belongs to an answer span:
$$p_{C_i} = softmax(W_{lin} \cdot MHA(z^*_{Q_i}, z_{C_i}, z_{C_i}) + b_{lin})$$
This probability is represented by $p_{C_i}$ for the $i^{th}$ context token. To measure the accuracy of span prediction, we compute sequence tagging loss employing cross-entropy loss:
$$L_{QASE} = -\frac{1}{N} \sum_{i=1}^N \sum_{j=0}^1 y_{ij}log(p_{C_{ij}})$$
where $j \in {0, 1}$ designates the classes \textbf{\textit{O}} and \textbf{\textit{I}}, and $y_{ij}$ is a binary indicator of whether the $i^{th}$ token is labeled as class $j$.

\subsection{Fine-Tuning and Inference}
\label{subsec:joint_finetune}

We fine-tune the PLMs employing a multi-task learning strategy that concurrently optimizes both the language modeling loss and the sequence tagging loss:
$$L = L_{LML} + \beta L_{QASE}$$
where $\beta$ is a hyper-parameter that determines the weight assigned to the span extraction task. This dual-objective approach substantially improves the PLMs' capability to generate contextually grounded and relevant answers. During the inference phase, only the generation component of the finely-tuned model is utilized.
\section{Experiments}
\label{sec:experiments}

This section presents the experimental framework, detailing the datasets used, experimental setup, comprehensive quantitative results of model performance, ablation studies, analysis of model factual consistency, and qualitative case studies.

\subsection{Datasets and Metrics}
\label{subsec:datasets}

We utilize three extractive MRC benchmark datasets:
\begin{itemize}[nosep,topsep=0pt]
    % \vspace{-0.2cm}
    \item[(1)] \textbf{SQuAD} \cite{rajpurkar-etal-2016-squad}: A benchmark dataset consisting of 100K+ questions with single-span answers. We use SQuAD v1.1. Since the official evaluation on v1.1 has long been ended, we report our results on the official v1.1 development set.
    % \vspace{-0.3cm}
    \item[(2)] \textbf{MultiSpanQA} \cite{li-etal-2022-multispanqa}: This dataset consists of over 6.5k question-answer pairs. Unlike most existing single-span answer MRC datasets, MultiSpanQA focuses on multi-span answers.
    % \vspace{-0.3cm}
    \item[(3)] \textbf{Quoref} \cite{dasigi-etal-2019-quoref}: A benchmark dataset containing more than 24K questions, with most answers being single-span and $\sim$10\% being multi-span.
    % \vspace{-0.3cm}
\end{itemize}

Following the conventions of the datasets' official leaderboards (listed in \ref{subsec:leaderboard}), we employ exact match (EM) and partial match (Overlap) F1 scores as metrics on MultiSpanQA, and exact match percentage and macro-averaged F1 score on SQuAD and Quoref.
\subsection{Experimental Setup}
\label{subsec:setup}

To assess the efficacy of the \textit{QASE} module independent of any specific language models, we conduct experiments with multiple open-source LLMs. Our tests include both decoder-only LLMs, such as Llama 2 \cite{touvron2023llama} and Alpaca \cite{alpaca}, and an encoder-decoder model family, Flan-T5 \cite{chung2022scaling}. For Llama 2 and Alpaca, we employ the pre-trained 7B version and fine-tune it using LoRA \cite{hu2021lora} combined with instruction-tuning (instruction templates are detailed in \ref{subsec:prompts}). For the Flan-T5 family, we fine-tune the small, base, and large versions. Detailed information about the trainable parameters for each model is provided in Table \ref{tab:params}.

\begin{table}[!htbp]
    \centering
    \small
    
    \begin{tabular}{c||ccc}
    \hline\hline
                           & \multicolumn{3}{c}{\textbf{Trainable Parameters}} \\
    \textbf{}              & no \textit{QASE}       & \textit{QASE}         & $\Delta$params       \\\hline\hline
    \textbf{\begin{tabular}[c]{@{}c@{}}Llama2/Alpaca\\with LoRA\end{tabular}} & 4.2M & 7.3M & 3.1M \\\hline
    \textbf{Flan-T5-Small} & 77.0M           & 78.2M          & 1.3M           \\\hline
    \textbf{Flan-T5-Base}  & 247.6M          & 248.9M         & 1.4M           \\\hline
    \textbf{Flan-T5-Large} & 783.2M          & 784.7M         & 1.5M           \\\hline\hline
    \end{tabular}
    \caption{Trainable parameters of experimented models.}
    \label{tab:params}
\end{table}

We determine the hyper-parameter $\beta = 1$ and the learning rate $lr = 1e-4$ using results from a grid search. For the LoRA fine-tuning of the Llama 2 and Alpaca models, we set a rank $r = 8$, $\alpha = 32$, and a dropout rate of $0.05$. The methodology for selecting these hyper-parameters is detailed in \ref{subsec:hp_selection}. All models are trained on individual GPUs with batch sizes ranging from 2 to 4, adjusted according to each GPU's VRAM capabilities. We employ four types of GPUs: A40, A10, A5500, and A100. Training continues for three epochs or until the models converge. Consistency is maintained across all variants of each base PLM in terms of GPU type, batch size, and training epochs.

\subsection{Does \textit{QASE} Mitigate Ill-Formed Generation?}
\label{subsec:results}

\begin{table*}[h]
    \centering
    \small
    \begin{tabular}{cr||c|c|c|c|c}
    \hline\hline
                                     & \textbf{} & \textbf{Llama2} & \textbf{Alpaca} & \textbf{Flan-T5-Small} & \textbf{Flan-T5-Base} & \textbf{Flan-T5-Large} \\\hline\hline
    \textbf{SQuAD}     & no \textit{QASE}   & 36.68 | 47.06          & 27.88 | 43.95          & 77.33 | 85.51          & 82.09 | 89.56          & 83.16 | 90.71          \\
    (EM | F1)                       & \textit{QASE} & \textbf{37.22} | \textbf{47.69} & \textbf{37.31} | \textbf{47.62} & \textbf{77.66} | \textbf{85.90} & \textbf{82.20} | \textbf{90.24} & \textbf{84.13} | \textbf{91.70} \\\hline
    \textbf{MultiSpanQA} & no \textit{QASE}   & 50.93 | 68.14          & \textbf{52.73} | 69.10          & \textbf{59.13} | 76.49          & 64.66 | 81.41          & \textbf{67.41} | 83.09          \\
    (EM F1 | Overlap F1)          & \textit{QASE}      & \textbf{51.75} | \textbf{70.39}   & 52.20 | \textbf{70.01}   & 59.08 | \textbf{77.10}      & \textbf{64.87} | \textbf{81.50}      & 66.92 | \textbf{84.22}      \\\hline
    \textbf{Quoref} & no \textit{QASE}        & 45.52 | 52.09   & 49.05 | 53.81               & 58.21 | 63.30      & 72.77 | 80.90      & 75.17 | 80.49      \\
    (EM | F1)                       & \textit{QASE}      & \textbf{54.28} | \textbf{60.44}   & \textbf{55.01} | \textbf{59.94}    & \textbf{60.70} | \textbf{66.88}      & \textbf{75.17} | \textbf{81.18}      & \textbf{76.19} | \textbf{82.13}   \\\hline\hline  
    \end{tabular}
    \caption{Performance (in \%) of fine-tuned PLMs with or without \textit{QASE} on each dataset.}
    \label{tab:main_results}
\end{table*}

To assess \textit{QASE} in mitigating ill-formed generation issue, we compare the performance of various PLMs fine-tuned with and without \textit{QASE}, as detailed in Table \ref{tab:main_results}. The conventional EM and partial match F1 scores effectively measure whether the generated answers match the gold answers in format on a token basis. Overall, models fine-tuned with \textit{QASE} consistently outperform those without it when measured by overlap F1 score. Specifically, for the SQuAD dataset, models with \textit{QASE} show an EM percentage increase of up to 33.8\% and an F1 score improvement of up to 8.4\% compared to vanilla fine-tuned models. For MultiSpanQA, improvements include up to 1.6\% in EM F1 and up to 3.3\% in overlap F1. Likewise, on the Quoref dataset, enhancements of up to 19.2\% in EM percentage and up to 16.0\% in F1 score are observed. These results confirm that \textit{QASE} enables generative-based PLMs to produce more accurate, contextually coherent, and higher-quality answers in MRC tasks compared to vanilla fine-tuning approaches. We also include discussions on performance discrepancies across different datasets and base PLMs in Appendix \ref{subsec:performance_discrepancy}.

For additional comparisons, we also evaluate the fine-tuned PLMs against their zero-shot performance, as outlined in Appendix \ref{subsec:full_results}. Specifically, on the SQuAD dataset, models using \textit{QASE} perform up to 5.6 times better in EM and 3.0 times better in F1 score compared to the zero-shot models. On the MultiSpanQA dataset, the EM improves by up to 124.4 times, and F1 score by up to 3.4 times. Similarly, on the Quoref dataset, the EM improves by up to 38.4 times, and F1 score by up to 11.2 times with \textit{QASE}. It is important to note that these substantial improvements stem from comparing zero-shot models to those fine-tuned with \textit{QASE}. Nonetheless, the previously discussed results comparing fine-tuned models with and without \textit{QASE} have clearly illustrated its effectiveness.

\subsubsection{\textit{QASE}-Enhaced PLMs vs SOTA LLMs and Extractive Approaches}
\label{subsec:model_compare}

Our top model, Flan-T5-Large$_{QASE}$, is further benchmarked against leading models on each dataset's official leaderboard, alongside zero-shot and few-shot GPT-3.5-Turbo and GPT-4. GPT-3.5-Turbo stands as one of OpenAI's most efficient models in terms of capability and cost, while GPT-4 shows superior reasoning abilities \cite{liu-etal-2023-system}. Studies indicate their superiority over traditional fine-tuning methods in most logical reasoning benchmarks \cite{liu2023evaluating}. The prompts used to query the GPT variants in zero-shot and few-shot scenarios are detailed in Appendix \ref{subsec:prompts}.

\begin{table}[h]
    \centering
    \small
    \begin{tabular}{l||cc}
    \hline\hline
    \textbf{}          & \textbf{EM}     & \textbf{F1} $\uparrow$     \\ \hline\hline
    GPT-3.5-Turbo      & 36.944          & 65.637          \\
    GPT-4              & 39.347          & 69.158          \\
    GPT-3.5-Turbo$_{2-shot}$ & 61.456 & 81.523 \\
    GPT-4$_{2-shot}$ & 74.096 & 88.216 \\
    Human \cite{rajpurkar-etal-2016-squad}  & 82.304          & 91.221          \\
    BERT-Large \cite{devlin-etal-2019-bert} & 84.328 & 91.281 \\
    MSRA NLNet (ensemble)   & \textbf{85.954} & 91.677      \\ \hline
    Flan-T5-Large$_{QASE}$ & 84.125 & \textbf{91.701} \\ \hline\hline
    \end{tabular}
    \caption{Flan-T5-Large$_{QASE}$ and baselines on \textbf{SQuAD}.}
    \label{tab:squad_results}
\end{table}
\textbf{On SQuAD}, as showed in Table \ref{tab:squad_results}, Flan-T5-Large$_{QASE}$ surpasses human performance \cite{rajpurkar-etal-2016-squad}, equaling the NLNet model from Microsoft Research Asia and the pre-trained BERT-Large \cite{devlin-etal-2019-bert}. %, which are ranked \#11 and \#13 on the v1.1 leaderboard respectively.%
Additionally, it surpasses two-shot GPT-4 by 13.6\% on EM and 4.0\% on F1.

\begin{table}[h]
    \centering
    \small
    \begin{tabular}{l||cc}
    \hline\hline
    \textbf{}                   & \textbf{EM F1}  & \textbf{Overlap F1} $\uparrow$ \\ \hline\hline
    GPT-3.5-Turbo$_{2-shot}$    & 52.987          & 78.588              \\
    GPT-3.5-Turbo               & 59.766          & 81.866              \\
    GPT-4                       & 64.027          & 82.731              \\ 
    LIQUID \cite{lee2023liquid} & \textbf{73.130} & 83.360              \\ 
    GPT-4$_{2-shot}$            & 65.399          & 83.546              \\ \hline
    Flan-T5-Large$_{QASE}$      & 66.918          & \textbf{84.221}     \\ \hline\hline
    \end{tabular}
    \caption{Performance of Flan-T5-Large$_{QASE}$ and baselines on \textbf{MultiSpanQA}.}
    \label{tab:multispanqa_results}
\end{table}
\textbf{On MultiSpanQA}, Table \ref{tab:multispanqa_results} shows that Flan-T5-Large$_{QASE}$ outperforms LIQUID \cite{lee2023liquid}, which currently ranks \#1 on the leaderboard, with respect to the overlap F1 score. Moreover, it surpasses zero-shot GPT-4 by 4.5\% on the exact match F1 and 1.5\% on the overlap F1, and two-shot GPT-4 by 2.3\% on the exact match F1 and 0.8\% on the overlap F1.

\begin{table}[h]
    \centering
    \small
    \begin{tabular}{l||cc}
    \hline\hline
    \textbf{}          & \textbf{EM}    & \textbf{F1} $\uparrow$    \\ \hline\hline
    GPT-3.5-Turbo      & 50.22          & 59.51          \\
    GPT-3.5-Turbo$_{2-shot}$ & 64.53 & 73.40 \\
    GPT-4              & 68.07          & 78.34          \\
    GPT-4$_{2-shot}$ & 74.36 & 80.15 \\
    CorefRoberta-Large \cite{ye-etal-2020-coreferential} & 75.80  & \textbf{82.81} \\ \hline
    Flan-T5-Large$_{QASE}$ & \textbf{76.19} & 82.13          \\ \hline\hline
    \end{tabular}
    \caption{Performance of Flan-T5-Large$_{QASE}$ and baselines on \textbf{Quoref}.}
    \label{tab:quoref_results}
\end{table}
\textbf{On Quoref}, Table \ref{tab:quoref_results} shows that Flan-T5-Large$_{QASE}$ is comparable to CorefRoberta-Large \cite{ye-etal-2020-coreferential}, which ranks \#9 on the leaderboard, with a 0.5\% higher exact match. Furthermore, it outperforms zero-shot GPT-4 by 11.9\% on EM and 4.8\% on F1, and two-shot GPT-4 by 2.5\% on both EM and F1.

All top-performing models on these datasets' leaderboards, equaling or exceeding Flan-T5-Large$_{QASE}$, are encoder-only extractive models. Therefore, these results demonstrate that \textit{QASE} shortens or closes the gap between generative and extractive approaches, enhancing PLMs to match the capabilities of SOTA extractive models and outperform leading LLMs on extractive MRC.
\subsection{Does \textit{QASE} Improve Factual Consistency?}
\label{subsec:q2}

While token-based EM and F1 scores measure the structural quality of generated text, they do not reflect factual accuracy relative to the context. To address this, we used $Q^2$ \cite{honovich2021q}, an automatic metric for assessing factual consistency in generated text, which uses question generation and answering methods over token-based matching. We compared fine-tuned Flan-T5-Large with and without \textit{QASE} in both single-span (SQuAD) and multi-span (MultiSpanQA) answer settings. Table \ref{tab:q2} shows that \textit{QASE}-enhanced models consistently outperform the vanilla fine-tuned model. On SQuAD, $Q^2$ NLI score is improved by 1.0\%, and on MultiSpanQA, it is improved by 16.0\%. 
% Beyond the $Q^2$ statistical analysis, our detailed case studies in Appendix \ref{subsec:case_studies} highlight Flan-T5-Large$_{QASE}$'s improved performance. These examples show the model's better alignment with relevant context, its enhanced understanding of complex sentences, its skill in synthesizing answers from dispersed information, and its superior use of pre-existing real-world knowledge in generating answers.

\begin{table}[!htbp]
    \centering
    \small
    \begin{tabular}{cr||cc}
    \hline\hline
                                          & Flan-T5-Large& $\boldsymbol{Q^2}$ \textbf{F1}  & $\boldsymbol{Q^2}$ \textbf{NLI} \\\hline\hline
    \multirow{2}{*}{\textbf{SQuAD}}       & no \textit{QASE}& 42.927          & 44.983          \\
                                          & \textit{QASE}& \textbf{43.624} & \textbf{45.419} \\\hline
    \multirow{2}{*}{\textbf{MultiSpanQA}} & no \textit{QASE}& 32.889          & 31.433          \\
                                          & \textit{QASE}& \textbf{34.732} & \textbf{36.452} \\\hline\hline
    % \multirow{2}{*}{\textbf{Quoref}}      & no \textit{QASE}& \textbf{33.588} & \textbf{33.950} \\
    %                                       & \textit{QASE}& 33.244          & 33.587 \\\hline\hline       
    \end{tabular}
    \caption{$Q^2$ scores of fine-tuned Flan-T5-Large with or without $QASE$ on each dataset.}
    \label{tab:q2}
\end{table}

\subsection{Computational Cost}

To assess the computational cost associated with \textit{QASE}, Table \ref{tab:params} reveals that incorporating the \textit{QASE} module incurs only a slight increase in the number of trainable parameters in PLMs. The degree of this increase varies based on the hidden sizes of the models. Remarkably, for the largest model, Flan-T5-Large, the addition of \textit{QASE} accounts for merely an extra 0.2\% in parameters. This underscores the fact that \textit{QASE} can substantially boost the performance of fine-tuned PLMs in MRC tasks without requiring significant additional computational resources.
\subsection{Ablation Studies}
\label{subsec:ablation}

We conduct ablation studies to assess the effectiveness of the \textit{QASE} architecture and to determine the optimal prompting strategy. Specifically, we compare Flan-T5-Large$_{QASE}$ with both the vanilla fine-tuned Flan-T5-Large$_{FT}$ and the baseline Flan-T5-Large$_{baseline}$. As shown in Figure \ref{fig:baseline} in Appendix \ref{subsec:ablation_details}, the baseline span extraction module does not include the \textbf{\textit{MHA}} component, rendering it a conventional architecture for fine-tuning pre-trained encoders on downstream sequence tagging tasks. For each configuration -- Flan-T5-Large$_{FT}$, Flan-T5-Large$_{QASE}$, and Flan-T5-Large$_{baseline}$ -- we explored both a question-first (\textit{qf}) and a context-first prompting strategy, with a detailed description of these strategies provided in Appendix \ref{subsec:ablation_details}.

Table \ref{tab:ablation} shows that the baseline-embedded model performs better with a question-first prompting strategy, as Flan-T5-Large$_{baseline_{qf}}$ surpasses Flan-T5-Large$_{baseline}$ and Flan-T5-Large$_{FT_{qf}}$. Conversely, the baseline span extraction module decreases performance in context-first prompting, where Flan-T5-Large$_{baseline}$ underperforms compared to Flan-T5-Large$_{FT}$. This suggests that adding an auxiliary span extraction module without careful design can negatively affect instruction fine-tuning. Meanwhile, the \textit{QASE}-enhanced model excels over both vanilla fine-tuned and baseline-embedded models in both prompting scenarios, demonstrating its architectural superiority. Specifically, in context-first setting, Flan-T5-Large$_{QASE}$ significantly outperforms Flan-T5-Large$_{baseline}$ with a 4.3\% higher F1. 

% Table \ref{tab:ablation} shows that both \textit{QASE}-enhanced models surpass their vanilla fine-tuned and baseline counterparts. Specifically, in context-first setting, Flan-T5-Large$_{QASE}$ significantly outperforms Flan-T5-Large$_{baseline}$ with a 5.3\% higher EM and 4.3\% higher F1. 

\begin{table}[h]
    \centering
    \small
    \begin{tabular}{c||cc}
    \hline\hline
             & \textbf{EM}     & \textbf{F1} $\uparrow$    \\\hline\hline
    Flan-T5-Large$_{baseline}$ & 79.877 & 87.918\\
    Flan-T5-Large$_{FT_{qf}}$  & 80.378 & 88.176\\
    Flan-T5-Large$_{baseline_{qf}}$ & 81.125          & 89.043          \\
    Flan-T5-Large$_{QASE_{qf}}$     & 81.485          & 89.077          \\
    Flan-T5-Large$_{FT}$ & 83.159 & 90.712 \\\hline
    Flan-T5-Large$_{QASE}$         & \textbf{84.125} & \textbf{91.701}  \\\hline\hline
    \end{tabular}
    \caption{Performance of vanilla, baseline-, and \textit{QASE}-enhanced fine-tuned Flan-T5-Large on \textbf{SQuAD}.}
    \label{tab:ablation}
\end{table}
\subsection{Qualitative Case Studies}
\label{subsec:case_studies}

In addition to the $Q^2$ statistical analysis in Section \ref{subsec:q2}, we also perform qualitative case studies to further demonstrate the effectiveness of \textit{QASE} in generating factual consistent answers.

\paragraph{Question Attended Alignment} Table \ref{tab:showcase_attention} showcases that Flan-T5-Large$_{QASE}$ more accurately identifies the key focus of the question and locates the pertinent factual information within the context, with the aid of the \textit{QASE} module. For instance, in \textbf{Sample 1}, Flan-T5-Large$_{QASE}$ correctly interprets the question as seeking the age difference between Newton and Manning, rather than the age of either individual, and accordingly provides the accurate answer. In contrast, Flan-T5-Large$_{FT}$ mistakenly provides Newton's age as the answer. Similarly, in \textbf{Sample 2}, Flan-T5-Large$_{QASE}$ accurately discerns that the question pertains to Thoreau's claim regarding the majority, generating in the correct answer, whereas Flan-T5-Large$_{FT}$ misguidedly responds with Thoreau's political philosophy.

\paragraph{Multi-Span Answers} Flan-T5-Large$_{QASE}$ also shows a notable improvement in comprehending complex, lengthy sentences and synthesizing answers from information that is sparsely distributed across multiple spans requiring logical processing. This capability is particularly valuable when the answer to a question does not directly stem from a single phrase. Table \ref{tab:showcase_multispan} provides examples of such instances. In \textbf{Sample 3}, the model needs to recognize that ESPN Deportes is the exclusive broadcaster in Spanish and that CBS, although mentioned, does not offer Spanish-language broadcasting. Combining these facts leads to the correct answer, that ESPN Deportes is the network that broadcast the game in Spanish. Flan-T5-Large$_{QASE}$ accurately generates this answer, whereas Flan-T5-Large$_{FT}$ incorrectly answers with ``CBS,'' likely due to confusion caused by the complex sentence structures and dispersed information. Similarly, in \textbf{Sample 4}, Flan-T5-Large$_{QASE}$ correctly identifies the question as seeking the name of the force related to a potential field between two locations. It successfully locates the relevant long sentence, deconstructs, and comprehends it to produce the correct answer, in contrast to Flan-T5-Large$_{FT}$, which incorrectly selects the first phrase mentioning ``force.'' In \textbf{Sample 5}, the question asks for the class most commonly not ascribed to the graph isomorphism problem. The model needs to deduce from the context that ``it is widely believed that the polynomial hierarchy does not collapse to any finite level,'' implying ``graph isomorphism is not NP-complete.'' Once again, Flan-T5-Large$_{QASE}$ arrives at the correct conclusion, while Flan-T5-Large$_{FT}$ does not.
\begin{table}[H]
    \centering
    \small
    \begin{tabular}{c||p{1.5in}}
        \hline\hline
        \multicolumn{2}{c}{\textbf{Sample 1}} \\
        \multicolumn{2}{p{2.8in}}{\textbf{Context:} This was the first Super Bowl to feature a quarterback on both teams who was the \#1 pick in their draft classes. Manning was the \#1 selection of the 1998 NFL draft, while Newton was picked first in 2011. The matchup also pits the top two picks of the 2011 draft against each other: Newton for Carolina and Von Miller for Denver. Manning and Newton also set the record for the largest \hl{age difference} between opposing Super Bowl quarterbacks at 13 years and 48 days (Manning was 39, Newton was \textcolor{red!90!green}{26}).} \\
        \multicolumn{2}{p{2.8in}}{\textbf{Question:} What was the \hl{age difference} between Newton and Manning in Super Bowl 50?} \\\hline
        \multicolumn{2}{p{2.8in}}{\textcolor{green!75!blue}{\textbf{Gold Answer:} 13 years and 48 days}} \\\hline\hline
        \begin{tabular}[c]{@{}c@{}}\textbf{Flan-T5-Large}$\boldsymbol{_{QASE}}$\\\textbf{Generation}\end{tabular} & \textcolor{green!75!blue}{13 years and 48 days} \\\hline
        \begin{tabular}[c]{@{}c@{}}Flan-T5-Large$_{FT}$\\Generation\end{tabular} & \textcolor{red!90!green}{26} \\\hline\hline
        \multicolumn{2}{c}{} \\\hline\hline

        \multicolumn{2}{c}{\textbf{Sample 2}} \\
        \multicolumn{2}{p{2.8in}}{\textbf{Context:} However, this definition is disputed by Thoreau's political philosophy, which \textcolor{red!90!green}{contrasts the conscience with the collective}. The individual is the ultimate arbiter of right and wrong. Beyond this, since only individuals act, only they can commit injustices. ... Thoreau acknowledges that the government may represent the will of the majority but it might also merely reflect the desires of elite politicians. Even a good government is "liable to be abused and perverted before the people can act through it." Furthermore, even if a government did express the voice of the people, this fact would not obligate the obedience of individuals who dissent. \hl{The majority may be powerful but it is not necessarily right.} What, then, is the appropriate relationship between the individual and the government?} \\
        \multicolumn{2}{p{2.8in}}{\textbf{Question:} What did Thoreau claim about \hl{the majority?}} \\\hline
        \multicolumn{2}{p{2.8in}}{\textcolor{green!75!blue}{\textbf{Gold Answer:} not necessarily right}} \\\hline\hline
        \begin{tabular}[c]{@{}c@{}}\textbf{Flan-T5-Large}$\boldsymbol{_{QASE}}$\\\textbf{Generation}\end{tabular} & it is \textcolor{green!75!blue}{not necessarily right} \\\hline
        \begin{tabular}[c]{@{}c@{}}Flan-T5-Large$_{FT}$\\Generation\end{tabular} & \textcolor{red!90!green}{conscience vs. the collective} \\\hline\hline
    \end{tabular}
    \caption{Comparisons of model attention alignment with question key aspects and relevant factual context between Flan-T5-Large$_{QASE}$ and Flan-T5-Large$_{FT}$.}
    \label{tab:showcase_attention}
    % \vspace{-0.5mm}
\end{table}
\begin{table}[H]
    \centering
    \small
    \begin{tabular}{c||p{1.5in}}
        \hline\hline
        \multicolumn{2}{c}{\textbf{Sample 3}} \\
        \multicolumn{2}{p{2.8in}}{\textbf{Context:} On December 28, 2015, \hl{ESPN Deportes} announced that they had reached an agreement with CBS and the NFL to be \hl{the exclusive Spanish-language broadcaster} of the game, marking the third dedicated Spanish-language broadcast of the Super Bowl. Unlike NBC and Fox, \hl{CBS does not have a Spanish-language outlet of its own} that could broadcast the game (though per league policy, a separate Spanish play-by-play call was carried on CBS's second audio program channel for over-the-air viewers). ...} \\
        \multicolumn{2}{p{2.8in}}{\textbf{Question:} Which network broadcast the game \hl{in Spanish?}} \\\hline
        \multicolumn{2}{p{2.8in}}{\textcolor{green!75!blue}{\textbf{Gold Answer:} ESPN Deportes}} \\\hline\hline
        \begin{tabular}[c]{@{}c@{}}\textbf{Flan-T5-Large}$\boldsymbol{_{QASE}}$\\\textbf{Generation}\end{tabular} & \textcolor{green!75!blue}{ESPN Deportes} \\\hline
        \begin{tabular}[c]{@{}c@{}}Flan-T5-Large$_{FT}$\\Generation\end{tabular} & \textcolor{red!90!green}{CBS} \\\hline\hline
        \multicolumn{2}{c}{} \\\hline\hline
        
        \multicolumn{2}{c}{\textbf{Sample 4}} \\
        \multicolumn{2}{p{2.8in}}{\textbf{Context:} A \textcolor{red!90!green}{conservative force} that acts on a closed system has an associated mechanical work that allows energy to convert only between kinetic or potential forms. This means that for a closed system, the net mechanical energy is conserved whenever a conservative force acts on the system. \hl{The force}, therefore, is \hl{related directly to the difference in potential energy between two different locations} in space, and \hl{can be considered to be an artifact} of the potential field in the same way that the direction and amount of a flow of water can be considered to be an artifact of the contour map of the elevation of an area.} \\
        \multicolumn{2}{p{2.8in}}{\textbf{Question:} What is \hl{the force} called \hl{regarding a potential field between two locations}?} \\\hline
        \multicolumn{2}{p{2.8in}}{\textcolor{green!75!blue}{\textbf{Gold Answer:} an artifact}} \\\hline\hline
        \begin{tabular}[c]{@{}c@{}}\textbf{Flan-T5-Large}$\boldsymbol{_{QASE}}$\\\textbf{Generation}\end{tabular} & \textcolor{green!75!blue}{an artifact} \\\hline
        \begin{tabular}[c]{@{}c@{}}Flan-T5-Large$_{FT}$\\Generation\end{tabular} & \textcolor{red!90!green}{conservative force} \\\hline\hline

        \multicolumn{2}{c}{} \\\hline\hline
        
        \multicolumn{2}{c}{\textbf{Sample 5}} \\
        \multicolumn{2}{p{2.8in}}{\textbf{Context:} The graph isomorphism problem is the computational problem of determining whether two finite graphs are isomorphic. An important unsolved problem in complexity theory is whether the graph isomorphism problem is in P, NP-complete, or \textcolor{red!90!green}{NP-intermediate}. The answer is not known, but \hl{it is believed that the problem is at least not NP-complete.} If graph isomorphism is NP-complete, the polynomial time hierarchy collapses to its second level. Since \hl{it is widely believed that the polynomial hierarchy does not collapse to any finite level}, it is believed that \hl{graph isomorphism is not NP-complete.} The best algorithm for this problem, due to Laszlo Babai and Eugene Luks has run time $2O(\sqrt{n log(n)})$ for graphs with n vertices.} \\
        \multicolumn{2}{p{2.8in}}{\textbf{Question:} What class is \hl{most commonly not ascribed to the graph isomorphism problem} in spite of definitive determination?} \\\hline
        \multicolumn{2}{p{2.8in}}{\textcolor{green!75!blue}{\textbf{Gold Answer:} NP-complete}} \\\hline\hline
        \begin{tabular}[c]{@{}c@{}}\textbf{Flan-T5-Large}$\boldsymbol{_{QASE}}$\\\textbf{Generation}\end{tabular} & \textcolor{green!75!blue}{NP-complete} \\\hline
        \begin{tabular}[c]{@{}c@{}}Flan-T5-Large$_{FT}$\\Generation\end{tabular} & \textcolor{red!90!green}{NP-intermediate} \\\hline\hline
    \end{tabular}
    \caption{Comparison of Flan-T5-Large$_{QASE}$ and Flan-T5-Large$_{FT}$ in understanding complex sentence structures.}
    \label{tab:showcase_multispan}
\end{table}

\paragraph{Real-World Knowledge} While our primary evaluation focuses on the model's proficiency in deriving answers from provided contexts, we also note that \textit{QASE} enhances the model's capacity to leverage real-world knowledge acquired during its pre-training phase. This improvement is attributed to \textit{QASE}'s ability to better align the model's focus on parts of the context that are relevant to the questions asked. Table \ref{tab:showcase_knowledge} presents an example of this phenomenon. In \textbf{Sample 6}, when asked about the California venue considered for the Super Bowl, Flan-T5-Large$_{QASE}$ correctly associates the San Francisco Bay Area with California, thus producing the accurate answer. On the other hand, Flan-T5-Large$_{FT}$ erroneously identifies a stadium in Miami as the answer. This example illustrates how \textit{QASE} not only improves context-based answer generation but also the model's application of pre-existing real-world knowledge to the questions posed.
\begin{table}[H]
    \centering
    \small
    \begin{tabular}{c||p{1.5in}}
        \hline\hline
        \multicolumn{2}{c}{\textbf{Sample 6}} \\
        \multicolumn{2}{p{2.8in}}{\textbf{Context:} The league eventually narrowed the bids to three sites: New Orleans' Mercedes-Benz Superdome, Miami's \textcolor{red!90!green}{Sun Life Stadium}, and the \hl{San Francisco Bay Area's} Levi's Stadium.} \\
        \multicolumn{2}{p{2.8in}}{\textbf{Question:} Which \hl{California} venue was one of three considered for Super Bowl 50?} \\\hline
        \multicolumn{2}{p{2.8in}}{\textcolor{green!75!blue}{\textbf{Gold Answer:} San Francisco Bay Area's Levi's Stadium}} \\\hline\hline
        \begin{tabular}[c]{@{}c@{}}\textbf{Flan-T5-Large}$\boldsymbol{_{QASE}}$\\\textbf{Generation}\end{tabular} & \textcolor{green!75!blue}{San Francisco Bay Area's Levi's Stadium} \\\hline
        \begin{tabular}[c]{@{}c@{}}Flan-T5-Large$_{FT}$\\Generation\end{tabular} & \textcolor{red!90!green}{Sun Life Stadium} \\\hline\hline
    \end{tabular}
    \caption{Comparison of Flan-T5-Large$_{QASE}$ and Flan-T5-Large$_{FT}$ in utilizing real-world knowledge.}
    \label{tab:showcase_knowledge}
    % \vspace{-0.5cm}
\end{table}

\paragraph{Common Failure Cases}
We observe that a recurring error made by Flan-T5-Large$_{QASE}$ is its \textbf{\textit{inability to correctly interpret Roman numerals}}, as seen in \textbf{Failure Sample 1} in Table \ref{tab:showcase_failure_short}, where the model is asked about the last Super Bowl held in California before Super Bowl 50. The correct answer, ``Super Bowl XXXVII,'' is clearly mentioned in the context, but Flan-T5-Large$_{QASE}$ incorrectly identifies ``Super Bowl XIX.'' The struggle with Roman numeral interpretation leads to errors even when the information is explicitly provided. 

Additionally, Flan-T5-Large$_{QASE}$ sometimes generates \textbf{\textit{slightly redundant phrases}}, though the excess is minimal. We argue that the generated answers are still correct in the given contexts, and the only reason they are not marked as 100\% accurate is due to the dataset's annotation scheme. For example, in \textbf{Failure Sample 2} in Table \ref{tab:showcase_failure_short}, Flan-T5-Large$_{QASE}$ accurately identifies that the ``entrance to studio 5'' is the critical element, whereas Flan-T5-Large$_{FT}$ simplifies the answer to just ``studio 5,'' missing the nuance. The inclusion of extra words highlights the model's attention to detail, even though it results in slight deviations from the gold-standard annotations. 

We observe \textbf{\textit{minor variations}} such as the omission or addition of non-essential words like ``the'' or punctuation marks. While technically considered ``mistakes'' in strict dataset annotations, these deviations do not affect the semantic accuracy of the model's responses. These are surface-level discrepancies rather than true comprehension errors. Such differences arguably shouldn't penalize the model, as they fall within acceptable linguistic flexibility. This prompts a discussion on evaluating model performance, where strict token-level matching may overlook underlying comprehension. For more examples and further discussion, see Appendix \ref{subsec:error_analysis}.

\begin{table}[H]
    \centering
    \small
    \begin{tabular}{c||p{1.5in}}
        \hline\hline
        \multicolumn{2}{c}{\textbf{Failure Sample 1}} \\
        \multicolumn{2}{p{2.8in}}{\textbf{Context:} On May 21, 2013, NFL owners at their spring meetings in Boston voted and awarded the game to Levi's Stadium. It is the first Super Bowl held in the San Francisco Bay Area since \textcolor{red!90!green}{Super Bowl XIX} in 1985, and the first in California since \hl{Super Bowl XXXVII} took place in San Diego in 2003.} \\
        \multicolumn{2}{p{2.8in}}{\textbf{Question:} Prior to Super Bowl 50, what was \hl{the last Super Bowl in California}?} \\\hline
        \multicolumn{2}{p{2.8in}}{\textcolor{green!75!blue}{\textbf{Gold Answer:} Super Bowl XXXVII}} \\\hline\hline
        \begin{tabular}[c]{@{}c@{}}\textbf{Flan-T5-Large}$\boldsymbol{_{QASE}}$\\\textbf{Generation}\end{tabular} & \textcolor{red!90!green}{Super Bowl XIX} \\\hline
        \begin{tabular}[c]{@{}c@{}}Flan-T5-Large$_{FT}$\\Generation\end{tabular} & \textcolor{red!90!green}{2010} \\\hline\hline
        \multicolumn{2}{c}{} \\\hline\hline
        
        \multicolumn{2}{c}{\textbf{Failure Sample 2}} \\
        \multicolumn{2}{p{2.8in}}{\textbf{Context:} ITV Tyne Tees was based at City Road for over 40 years after its launch in January 1959. In 2005 it moved to a new facility on The Watermark business park next to the MetroCentre in Gateshead. \hl{The entrance to studio 5} at the City Road complex gave its name to the 1980s music television programme, The Tube. ...} \\
        \multicolumn{2}{p{2.8in}}{\textbf{Question:} What gave its name to the 1980s music television program ``The Tube''?} \\\hline
        \multicolumn{2}{p{2.8in}}{\textcolor{green!75!blue}{\textbf{Gold Answer:} The entrance to studio 5}} \\\hline\hline
        \begin{tabular}[c]{@{}c@{}}\textbf{Flan-T5-Large}$\boldsymbol{_{QASE}}$\\\textbf{Generation}\end{tabular} & \textcolor{green!75!blue}{entrance to studio 5} \textcolor{red!90!green}{at the City Road complex} \\\hline
        \begin{tabular}[c]{@{}c@{}}Flan-T5-Large$_{FT}$\\Generation\end{tabular} & \textcolor{red!90!green}{studio 5} \\\hline\hline
    \end{tabular}
    \caption{Failure cases of Flan-T5-Large$_{QASE}$.}
    \label{tab:showcase_failure_short}
    % \vspace{-0.5mm}
\end{table}
\section{Discussions}
\label{sec:discussions}
In this section, we briefly address the weak performance of Flan-T5 zero-shot and Llama 2 on extractive MRC tasks, despite their strong language understanding abilities. We note that a comprehensive analysis is beyond our study's scope. Our goal is to gain insights into further improving these PLMs' effectiveness in extractive MRC.

\subsection{Flan-T5 Zero-Shot Performance}
\label{subsec:flan-t5_zeroshot_main}
Despite being trained on SQuAD during pre-training, zero-shot Flan-T5 models demonstrate poor performance across datasets, including SQuAD. While a comprehensive analysis of Flan-T5's performance is beyond the focus of our study, we briefly explore potential reasons for this underperformance to gain better insights. This underperformance may stem from their training on a wide range of tasks (1,836 tasks), focusing on free-form generation, QA, and reasoning tasks, rather than being finely optimized for extractive QA tasks like MRC. Additionally, generative models like Flan-T5 and Llama 2 generally struggle in MRC tasks, as discussed earlier. For extended discussions, refer to Appendix \ref{subsec:flan-t5_zeroshot}.

For fairness in our zero-shot experiments, we compare our prompt template with Google's instruct-tuning prompts for Flan-T5 on the SQuAD v1 dataset. Our results, as illustrated in Table \ref{tab:squad_zeroshot_prompts}, reveal that our prompt template achieves the highest F1 score. This implies that Flan-T5's lower zero-shot performance on MRC is expected.

\subsection{Llama 2 Performance}
\label{subsec:llama_performance_main}
We also observe that models based on Llama 2 and Alpaca consistently underperform compared to those based on Flan-T5, across zero-shot and fine-tuned scenarios, with or without \textit{QASE}. This discrepancy may arise from the significant difference in the number of trainable parameters, as illustrated in Table \ref{tab:params}, during fine-tuning. Additionally, factors such as differences in pre-training datasets and varied adaptation to tasks due to structural disparities can also contribute to this performance gap. While acknowledging these factors, conducting a comprehensive comparison of different generative model architectures in extractive MRC tasks exceeds the scope of our study. For further discussion, please refer to Appendix \ref{subsec:llama_performance}.

\subsection{Performance Discrepancy across Different Base PLMs and Datasets}
\label{subsec:performance_discrepancy_main}

While we observe significant performance improvements with \textit{QASE} across various base PLMs and datasets, the extent of the gains varies by dataset. Specifically, Quoref shows the largest improvement, as shown in Table \ref{tab:performance_discrepancy} in Appendix \ref{subsec:performance_discrepancy}, partly due to its weaker baselines. For instance, a Flan-T5-Small model fine-tuned without \textit{QASE} achieves F1 scores of 85.51\% on SQuAD, 76.49\% on MultiSpanQA, and 63.30\% on Quoref. Higher baseline scores on datasets such as SQuAD make further improvements more challenging but still notable.

Model-wise, Llama 2 and Alpaca show larger improvements than Flan-T5 models, with some exceptions on MultiSpanQA. This is likely due to Flan-T5's higher baseline performance, as discussed in Sections \ref{subsec:flan-t5_zeroshot_main}, \ref{subsec:llama_performance_main}, and Appendix \ref{subsec:flan-t5_zeroshot}, \ref{subsec:llama_performance}. While there are potential factors that explain some of Flan-T5’s superior performance in context-based question answering, a comprehensive comparison of model architectures in MRC tasks is beyond the scope of this study. For further discussion, refer to Appendix \ref{subsec:performance_discrepancy}.

% \subsection{Prompting Strategies}
% While we acknowledge the prevalent use of prompting strategies like in-context learning, our main focus is not on evaluating these strategies but rather on improving the performance of generative PLMs on MRC tasks through fine-tuning, with and without the integration of \textit{QASE}. However, we do experiment with various prompt templates, as discussed in Section \ref{subsec:ablation} and Appendix \ref{subsec:flan-t5_zeroshot}.
\section{Conclusions and Future Work}

In this study, we address \textit{out-of-control generation} issue of generative PLMs in extractive MRC using \textit{QASE}, a lightweight question-attended span extraction module, during the fine-tuning of PLMs. Our experiments show that \textit{QASE}-enhanced PLMs generate better-quality responses with improved formality and factual consistency, matching SOTA extractive models and outperforming few-shot GPT-4 by a significant margin on all three extractive MRC datasets, bridging the gap between generative and extractive models in extractive MRC tasks. Importantly, \textit{QASE} improves performance without a significant increase in computational costs, benefiting researchers with limited resources.

As the next step, we plan to conduct interpretability analyses to examine the performance discrepancies across different base PLMs and datasets.

In the future, we aim to evaluate our model on generative MRC tasks, such as \citet{nguyen2016ms}, to gauge its effectiveness in handling more intricate scenarios. Additionally, a significant emphasis will be placed on assessing the model's overall capability in answer generation, with a specific focus on human perception. This involves incorporating human annotators alongside automatic metrics. Looking further ahead, we aspire to extend our research to explore strategies for mitigating input- and context-conflicting hallucinations in LLMs.
\clearpage
\section*{Limitations}
\label{sec:limitations}
Due to our limited computational resources, we have been able to perform our experiments on models no larger than Flan-T5-Large. This same constraint leads us to only fine-tuning of Llama 2 and Alpaca with LoRA. We note that models based on Llama 2 and Alpaca generally underperform those based on Flan-T5. Apart from the inherent distinctions between decoder-only and encoder-decoder models, and their suitability for different tasks (as seen from the models' zero-shot performance), a possible factor could be the number of trainable parameters during fine-tuning. Specifically, fine-tuning Llama 2 and Alpaca with LoRA results in only 4.2M trainable parameters, while even the smallest Flan-T5 model provides 77.0M trainable parameters, as shown in Table \ref{tab:params}. We acknowledge that many researchers face similar computational resource limitations. Therefore, our research should be very useful, proposing this lightweight module capable of enhancing smaller PLMs to outperform leading LLMs on MRC tasks like these, achieving a balance of effectiveness and affordability.

One foreseeable limitation of our work is the dependency of the fine-tuning process on answer span annotations, since \textit{QASE} works as an auxiliary supervised span extraction module. This reliance on annotated data could potentially limit the model's broader applicability. A prospective exciting future direction to address this limitation is to develop a semi- or unsupervised module that focuses on selecting relevant spans or rationales within a given context. By integrating this module with our current model, we could significantly improve its generalization capabilities, thereby making it more adaptable and effective across a wider range of scenarios.

One popular method to enhance the formality of answers generated by LLMs is through prompt engineering, paired with few-shot or in-context learning techniques. While these strategies offer great advantages, our ultimate goal is to create a system with broad domain generalization, one that minimizes the need for extensive, calibrated prompt engineering and sample selections for task adaptation. Although developing a robust prompt engineering framework or paradigm is an appealing direction, our current focus diverges from this path. As a long-term goal, we aim for a solution that handles diverse tasks with minimal task-specific tuning.

\bibliography{filtered_anthology, custom}
\newpage
\appendix

\section{Detailed Experiment Setup and Results}
\label{sec:appendix}

\subsection{Dataset Leaderboard}
\label{subsec:leaderboard}

Below are the official leaderboards all the datasets we refer to:

\begin{table}[!htbp]
    \centering
    \small
    \begin{tabular}{lp{2in}}
    \hline\hline
    \textbf{SQuAD} & \url{https://rajpurkar.github.io/SQuAD-explorer/} \\\hline
    \textbf{MultiSpanQA} & \url{https://multi-span.github.io/} \\\hline
    \textbf{Quoref} & \url{https://leaderboard.allenai.org/quoref/submissions/public} \\\hline\hline
    \end{tabular}
    \caption{Dataset official leaderboards.}
    \label{tab:leaderboards}
\end{table}

\subsection{Hyper-Parameter Selection}
\label{subsec:hp_selection}
In this section, we outline the process for selecting the hyper-parameter $\beta$ and detail our approach to LoRA fine-tuning.

For selecting $\beta$, we use a grid search method, exploring values from 0.5 to 2 in increments of 0.1, on 30\% of the MultiSpanQA training dataset. This process leads to the determination that $\beta = 1$ empirically yield the best performance, hence it is selected for use in our experiments.

To select the learning rate $lr$, we conduct a grid search, testing values from $\{1e-5, 5e-5, 1e-4, 5e-4, 1e-3\}$ on 30\% of the MultiSpanQA training dataset. Empirically, the value $1e-4$ demonstrates the best performance and is therefore chosen for our experiments. This selection is in agreement with the default $lr$ value used in Meta's official Llama 2 fine-tuning recipe\footnote{\href{https://github.com/facebookresearch/llama-recipes/blob/main/src/llama_recipes/configs/training.py}{Link to the fine-tuning configuration of Meta's official Llama 2 recipe.}}.

In the case of LoRA fine-tuning, we follow the established methodology as outlined by \citet{hu2021lora}. This involves applying LoRA to Llama 2 and the pre-trained Alpaca models by freezing their pre-trained weights and integrating trainable rank decomposition matrices at every layer of their Transformer structures, aimed at reducing the number of trainable parameters to enhance computational efficiency. We implement this using the PEFT package\footnote{\href{https://github.com/huggingface/peft}{Link to the Hugging Face PEFT implementation.}}. The fine-tuning hyper-parameters for LoRA are set according to the default settings specified in Meta's official Llama 2 fine-tuning recipe\footnote{\href{src/llama_recipes/configs/peft.py}{Link to the LoRA hyper-parameter configuration of Meta's official Llama 2 recipe.}}, which include a rank $r = 8$, $\alpha = 32$, and a dropout rate of $0.05$.

\subsection{Full Experiment Results}
\label{subsec:full_results}
In addition to the highlighted results presented in Section \ref{sec:experiments}, we also compare the fine-tuned PLMs to their corresponding base PLMs in zero-shot settings. The results, presented in Table \ref{tab:plm_qase}, show that fine-tuning with \textit{QASE} improves performance across all datasets. Specifically, on the SQuAD dataset, models using \textit{QASE} perform up to 5.6 times better in exact match and 3.0 times better in F1 score compared to the original models. On the MultiSpanQA dataset, the exact match improves by up to 124.4 times, and F1 score by up to 3.4 times. Similarly, on the Quoref dataset, the exact match improves by up to 38.4 times, and F1 score by up to 11.2 times with \textit{QASE}.

\begin{table*}[!htbp]
    \centering
    \small
    \begin{tabular}{l||cc|cc|cc}
    \hline\hline
    \textbf{}          & \multicolumn{2}{c|}{\textbf{MultiSpanQA}} & \multicolumn{2}{c|}{\textbf{SQuAD}} & \multicolumn{2}{c}{\textbf{Quoref}} \\
    \textbf{}          & \textbf{EM F1}    & \textbf{Overlap F1}  & \textbf{EM}      & \textbf{F1}     & \textbf{EM}      & \textbf{F1}      \\ \hline\hline
    Llama2        & 7.354  & 34.031 & 13.443 & 28.931 & 5.02  & 28.91 \\
    Llama2$_{FT}$ & 50.934 & 68.140 & 36.679 & 47.055 & 45.52 & 52.09 \\
    Llama2$_{QASE}$       & \textbf{51.748}   & \textbf{70.389}      & \textbf{37.219}  & \textbf{47.686} & \textbf{54.28}   & \textbf{60.44}   \\ \hline
    Alpaca        & 15.201 & 42.759 & 18.259 & 33.871 & 9.67     & 30.02    \\
    Alpaca$_{FT}$ & \textbf{52.730} & 69.099 & 27.881 & 43.950 & 49.05 & 53.81 \\
    Alpaca$_{QASE}$        & 52.196   & \textbf{70.008}      & \textbf{37.313}  & \textbf{47.622} & \textbf{55.01}            & \textbf{59.94}                \\ \hline
    Flan-T5-Small & 0.475  & 22.539 & 13.878 & 28.710 & 1.58  & 5.96  \\
    Flan-T5-Small$_{FT}$ & \textbf{59.128} & 76.494 & 77.332 & 85.513 & 58.21 & 63.30 \\
    Flan-T5-Small$_{QASE}$ & 59.080   & \textbf{77.103}      & \textbf{77.663}  & \textbf{85.901} & \textbf{60.70}   & \textbf{66.88}   \\ \hline
    Flan-T5-Base  & 4.113  & 37.694 & 37.596 & 51.747 & 27.08 & 34.38 \\
    Flan-T5-Base$_{FT}$ & 64.659 & 81.408 & 82.090 & 89.558 & 72.77 & 80.90 \\
    Flan-T5-Base$_{QASE}$  & \textbf{64.874}   & \textbf{81.498}      & \textbf{82.204}  & \textbf{90.240} & \textbf{75.17}   & \textbf{81.18}   \\ \hline
    Flan-T5-Large & 13.907 & 51.501 & 16.149 & 37.691 & 15.96 & 24.10 \\
    Flan-T5-Large$_{FT}$ & \textbf{67.408} & 83.094 & 83.159 & 90.712 & 75.17 & 80.49 \\
    Flan-T5-Large$_{QASE}$ & 66.918   & \textbf{84.221}      & \textbf{84.125}  & \textbf{91.701} & \textbf{76.19}   & \textbf{82.13}   \\ \hline\hline
    \end{tabular}
    \caption{Performance of zero-shot PLMs and fined-tuned PLMs with and without \textit{QASE}.}
    \label{tab:plm_qase}
\end{table*}

\subsection{Instruction Templates and Model Prompts}
\label{subsec:prompts}
Table \ref{tab:prompts} provides the instruction and prompt templates used for fine-tuning the PLMs and for zero-shot and few-shot querying of PLMs and GPT variants across both single- and multi-span answer datasets. In few-shot prompting scenarios, examples are randomly selected from the training set.

\begin{table*}
    \centering
    \small
    \begin{tabular}{p{1.8in}|p{4.2in}}
        \hline\hline
        \textbf{Fine-tuning} PLMs & Instruction: Using the provided context, answer the question with exact phrases and avoid explanations.\newline - - - \newline Context: \{context\}\newline - - - \newline Question: \{question\}\newline - - -\newline Answer: \\\hline
        \textbf{Zero-shot} prompting PLMs and GPT variants on \textbf{single-span} answer dataset, SQuAD & Instruction: Using the provided context, answer the question with exact phrases and avoid explanations. \textit{[Format the response as follows: ["answer1", "answer2", ...].]}$^*$\newline - - - \newline Context: \{context\}\newline - - - \newline Question: \{question\}\newline - - -\newline Answer: \\\hline
         \textbf{Few-shot} prompting PLMs and GPT variants & Instruction: Using the provided context, answer the question with exact phrases and avoid explanations. \textit{[Format the response as follows: ["answer1", "answer2", ...].]}$^*$\newline\newline - - - \newline Example {i}\newline Context: \{example context\}\newline - - - \newline Question: \{example question\}\newline - - -\newline Answer: {example answer}\newline\newline - - -\newline Context: \{context\}\newline - - - \newline Question: \{question\}\newline - - -\newline Answer: \\\hline\hline
    \end{tabular}
    \caption{Templates for fine-tuning instructions and zero-shot and few-shot query prompts. $^*$Text in square bracket is only added for multi-span answer datasets, MultiSpanQA and Quoref.}
    \label{tab:prompts}
\end{table*}

\subsection{Ablation Studies Details}
\label{subsec:ablation_details}
\begin{figure}[!htbp]
  \centering
  \includegraphics[width=0.48\textwidth]{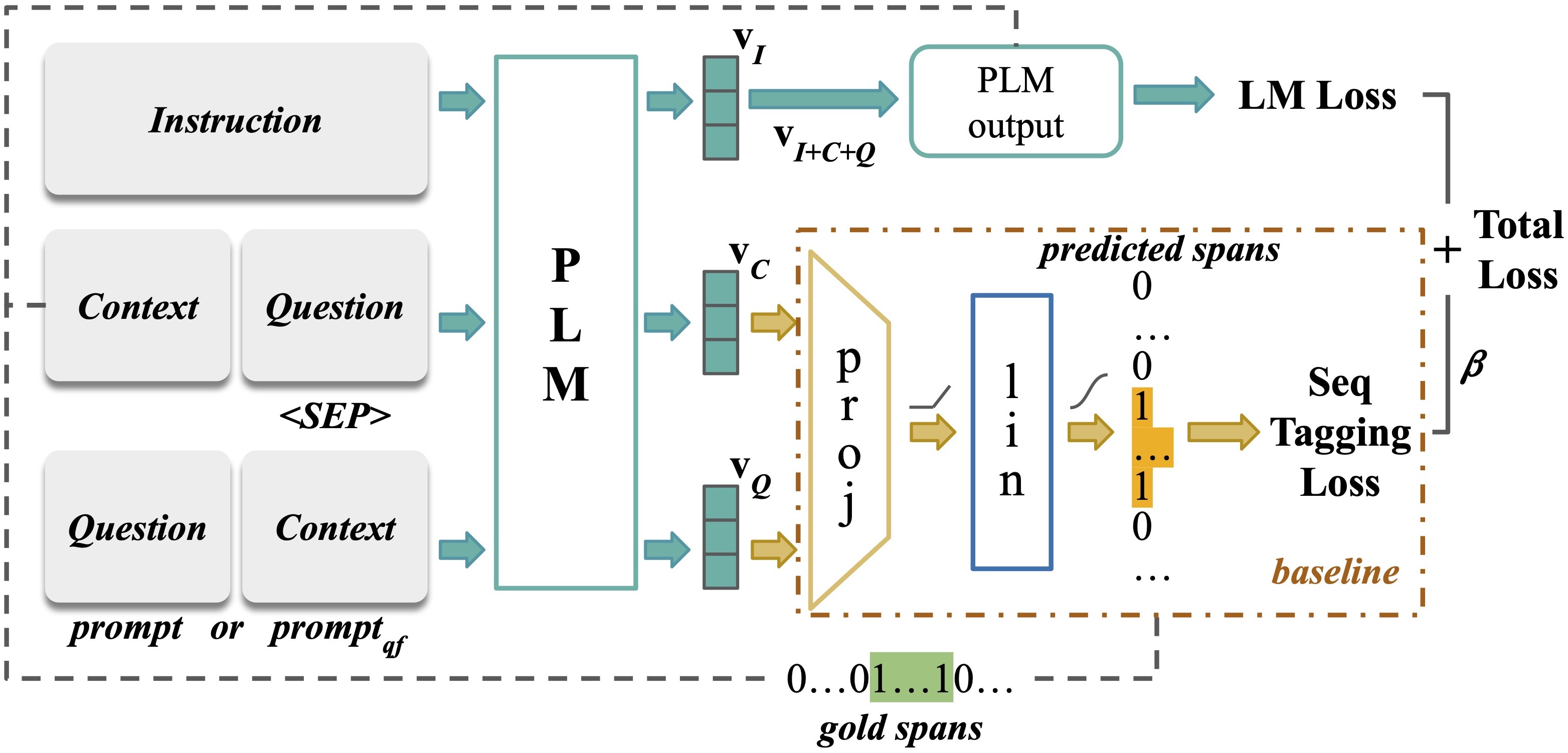}
  \caption{Baseline-embedded model architecture.}
  \label{fig:baseline}
\end{figure}

Figure \ref{fig:baseline} depicts the architecture of the model we use for the ablation studies, with a baseline span extraction module. The baseline span extraction module omits the \textit{MHA} component, typifying a standard architecture for fine-tuning pre-trained encoders for downstream sequence tagging tasks. It closely resembles the approach by \citet{xu2021attention}, with two key differences: \textbf{(a)} our baseline model integrates both query and context token embeddings to provide additional contextual information, and \textbf{(b)} instead of directly computing the extraction loss, our model includes additional projection and linear layers. The baseline-embedded Flan-T5-Large models are fine-tuned with the same configurations as Flan-T5-Large$_{QASE}$ including learning rate, weight decay, batch size, epoch number, and GPU type.

We experiment with 2 prompting strategies for ablation studies:
\begin{itemize}
    \item \textbf{Context-first prompting:} The default prompting strategy we utilize for fine-tuning PLMs, both with and without \textit{QASE}. In this setting, the prompt is ordered as "<instruction tokens> <context tokens> <question tokens>".
    \item \textbf{Question-first prompting (\textit{qf}):} Following BERT's standard fine-tuning procedures. In this setting, the prompt is ordered as "<instruction tokens> <question tokens> <SEP> <context tokens>". <SEP> is a special separator token.
\end{itemize}

\subsection{Qualitative Error Analysis}
\label{subsec:error_analysis}

\begin{table}[H]
    \centering
    \small
    \begin{tabular}{c||p{1.5in}}
        \hline\hline
        \multicolumn{2}{c}{\textbf{Failure Sample 1}} \\
        \multicolumn{2}{p{2.8in}}{\textbf{Context:} On May 21, 2013, NFL owners at their spring meetings in Boston voted and awarded the game to Levi's Stadium. It is the first Super Bowl held in the San Francisco Bay Area since \textcolor{red!90!green}{Super Bowl XIX} in 1985, and the first in California since \hl{Super Bowl XXXVII} took place in San Diego in 2003.} \\
        \multicolumn{2}{p{2.8in}}{\textbf{Question:} Prior to Super Bowl 50, what was \hl{the last Super Bowl in California}?} \\\hline
        \multicolumn{2}{p{2.8in}}{\textcolor{green!75!blue}{\textbf{Gold Answer:} Super Bowl XXXVII}} \\\hline\hline
        \begin{tabular}[c]{@{}c@{}}\textbf{Flan-T5-Large}$\boldsymbol{_{QASE}}$\\\textbf{Generation}\end{tabular} & \textcolor{red!90!green}{Super Bowl XIX} \\\hline
        \begin{tabular}[c]{@{}c@{}}Flan-T5-Large$_{FT}$\\Generation\end{tabular} & \textcolor{red!90!green}{2010} \\\hline\hline
        \multicolumn{2}{c}{} \\\hline\hline
        
        \multicolumn{2}{c}{\textbf{Failure Sample 2}} \\
        \multicolumn{2}{p{2.8in}}{\textbf{Context:} The \textcolor{red!90!green}{Super Bowl 50} halftime show was headlined by the British rock group Coldplay with special guest performers \hl{Beyonce} and Bruno Mars, who headlined the \hl{Super Bowl XLVII} and Super Bowl XLVIII halftime shows, respectively.} \\
        \multicolumn{2}{p{2.8in}}{\textbf{Question:} At which Super Bowl did \hl{Beyonce} headline the halftime show?} \\\hline
        \multicolumn{2}{p{2.8in}}{\textcolor{green!75!blue}{\textbf{Gold Answer:} Super Bowl XLVII}} \\\hline\hline
        \begin{tabular}[c]{@{}c@{}}\textbf{Flan-T5-Large}$\boldsymbol{_{QASE}}$\\\textbf{Generation}\end{tabular} & \textcolor{red!90!green}{Super Bowl 50} \\\hline
        \begin{tabular}[c]{@{}c@{}}Flan-T5-Large$_{FT}$\\Generation\end{tabular} & \textcolor{red!90!green}{Super Bowl 50} \\\hline\hline
    \end{tabular}
    \caption{Failure cases of Flan-T5-Large$_{QASE}$ in interpreting Roman numerals.}
    \label{tab:showcase_failure_roman}
    % \vspace{-0.5mm}
\end{table}
\paragraph{Challenges in Roman Numeral Interpretation} We observe that a recurring error made by Flan-T5-Large$_{QASE}$ is its inability to correctly interpret Roman numerals, as evidenced in Table \ref{tab:showcase_failure_roman}. In \textbf{Failure Sample 1}, the model is asked about the last Super Bowl held in California before Super Bowl 50. The correct answer, ``Super Bowl XXXVII,'' is clearly mentioned in the context, but Flan-T5-Large$_{QASE}$ incorrectly identifies ``Super Bowl XIX.'' Similarly, in \textbf{Failure Sample 2}, the context states that Beyonce headlined Super Bowl XLVII, yet the model incorrectly identifies ``Super Bowl 50'' as the answer, despite the clear mention of Super Bowl XLVII in the question and context. These examples indicate that Flan-T5-Large$_{QASE}$ struggles with Roman numeral interpretation, leading to errors even when the information is explicitly provided in the text.

\begin{table}[H]
    \centering
    \small
    \begin{tabular}{c||p{1.5in}}
        \hline\hline
        \multicolumn{2}{c}{\textbf{Failure Sample 3}} \\
        \multicolumn{2}{p{2.8in}}{\textbf{Context:} Super Bowl 50 was an American football game to determine the champion of the National Football League (NFL) for the 2015 season. The game was played on February 7, 2016, at Levi's Stadium in the \textcolor{red!90!green}{San Francisco Bay Area} at \hl{Santa Clara}, California. ...} \\
        \multicolumn{2}{p{2.8in}}{\textbf{Question:} What \hl{city} did Super Bowl 50 take place in?} \\\hline
        \multicolumn{2}{p{2.8in}}{\textcolor{green!75!blue}{\textbf{Gold Answer:} Santa Clara}} \\\hline\hline
        \begin{tabular}[c]{@{}c@{}}\textbf{Flan-T5-Large}$\boldsymbol{_{QASE}}$\\\textbf{Generation}\end{tabular} & \textcolor{green!75!blue}{Santa Clara}\textcolor{red!90!green}{, California} \\\hline
        \begin{tabular}[c]{@{}c@{}}Flan-T5-Large$_{FT}$\\Generation\end{tabular} & \textcolor{red!90!green}{San Francisco Bay Area} \\\hline\hline
        \multicolumn{2}{c}{} \\\hline\hline
        
        \multicolumn{2}{c}{\textbf{Failure Sample 4}} \\
        \multicolumn{2}{p{2.8in}}{\textbf{Context:} ITV Tyne Tees was based at City Road for over 40 years after its launch in January 1959. In 2005 it moved to a new facility on The Watermark business park next to the MetroCentre in Gateshead. \hl{The entrance to studio 5} at the City Road complex gave its name to the 1980s music television programme, The Tube. ...} \\
        \multicolumn{2}{p{2.8in}}{\textbf{Question:} What gave its name to the 1980s music television program ``The Tube''?} \\\hline
        \multicolumn{2}{p{2.8in}}{\textcolor{green!75!blue}{\textbf{Gold Answer:} The entrance to studio 5}} \\\hline\hline
        \begin{tabular}[c]{@{}c@{}}\textbf{Flan-T5-Large}$\boldsymbol{_{QASE}}$\\\textbf{Generation}\end{tabular} & \textcolor{green!75!blue}{entrance to studio 5} \textcolor{red!90!green}{at the City Road complex} \\\hline
        \begin{tabular}[c]{@{}c@{}}Flan-T5-Large$_{FT}$\\Generation\end{tabular} & \textcolor{red!90!green}{studio 5} \\\hline\hline
    \end{tabular}
    \caption{Failure cases of Flan-T5-Large$_{QASE}$ in generating redundant phrases.}
    \label{tab:showcase_failure_redundant}
    % \vspace{-0.5mm}
\end{table}
\paragraph{Minor Redundancies in Generation} Another common error we observe is that Flan-T5-Large$_{QASE}$ tends to generate slightly redundant phrases, though the excess is minimal. We argue that the generated answers are still correct in the given contexts, and the only reason they are not marked as 100\% accurate is due to the dataset's annotation scheme. In \textbf{Failure Sample 3}, for instance, Flan-T5-Large$_{FT}$ produces a completely incorrect answer, while Flan-T5-Large$_{QASE}$ provides the correct answer, but with a minor additional word, ``California,'' which does not detract from its correctness. Similarly, in \textbf{Failure Sample 4}, Flan-T5-Large$_{QASE}$ accurately identifies that the ``entrance to studio 5'' is the critical element, whereas Flan-T5-Large$_{FT}$ simplifies the answer to just ``studio 5,'' missing the nuance. The inclusion of extra words in Flan-T5-Large$_{QASE}$'s responses highlights the model's attention to detail, even though it results in slight deviations from the gold-standard annotations, as shown in Table \ref{tab:showcase_failure_redundant}.

\paragraph{Surface-Level Variations in Output} Another pattern we observe involves minor variations in Flan-T5-Large$_{QASE}$'s outputs, such as the omission or addition of non-essential words like ``the'' or punctuation marks, such as quotation marks. These deviations, while technically considered ``mistakes'' in the strict context of dataset annotations, do not alter the semantic accuracy of the model's responses. For example, if the model omits a definite article or does not replicate quotation marks around a phrase, the intended meaning of the answer remains intact. These variations are more reflective of surface-level discrepancies rather than true comprehension errors. One could argue that such differences should not penalize the model, as they fall within the bounds of acceptable linguistic flexibility. This raises an interesting discussion about how we evaluate model performance, particularly in cases where strict adherence to token-level matching might overlook the model’s underlying comprehension. These ``errors'' suggest that Flan-T5-Large$_{QASE}$ generates answers with a focus on meaning rather than perfect alignment with rigid annotation structures, thus offering a more human-like adaptability in its output.

\section{Extended Discussion on Model Performance}
In this section, we engage in a detailed discussion on the performance of the Flan-T5 family of models and Llama 2 in MRC tasks. Our aim is to gain insights into the reasons behind the modest zero-shot performance of these large PLMs on MRC tasks, despite their adeptness at handling other complex NLP tasks such as dialogue generation and summarization. Although a comprehensive analysis falls outside the scope of our current study, exploring these performance nuances can provide valuable perspectives on how to potentially enhance the effectiveness of these PLMs on similar tasks.

\subsection{Discussion on Flan-T5 Zero-Shot Performance}
\label{subsec:flan-t5_zeroshot}

We observe that the zero-shot performance of Flan-T5 models across all datasets, including SQuAD, remains low as shown in Table \ref{tab:plm_qase}, despite being instruct-tuned on the SQuAD dataset during the pre-training phase. This underperformance might stem from the fact that Flan-T5 models, although trained on the <SQuAD, Extractive QA> task, are also trained on a broad spectrum of 1,836 tasks, predominantly focusing on free-form generation, QA, and reasoning tasks \cite{chung2022scaling}. Consequently, these models are not finely optimized for extractive QA tasks like MRC, especially under metrics like exact match and F1, particularly for the smaller to larger variants under study. The larger XL and XXL variants may exhibit better performance in these tasks. Furthermore, as discussed in the previous sections, generative models, including Llama 2, Alpaca, and GPT variants, generally show limited effectiveness in MRC tasks in zero-shot settings, underscored by their poorer performance despite having significantly larger model parameters compared to the Flan-T5 variants we experiment with.

To ensure that our zero-shot experiment's prompts do not adversely affect Flan-T5's performance, we compare our prompt template, detailed in Table \ref{tab:prompts}, with those Google released for Flan-T5's instruct-tuning on the SQuAD v1 dataset\footnote{\href{https://github.com/google-research/FLAN/blob/main/flan/templates.py}{Link to Flan-T5 instruct-tuning prompt templates.}}. Our template, similar to Google's, differs mainly by including "with exact phrases and avoid explanations." This difference could potentially affect performance, yet our subsequent experiments demonstrate otherwise.

We conduct a series of experiments to assess the zero-shot performance of Flan-T5-Large on SQuAD, using Google released templates for Flan-T5 instruct-tuning. We select three templates of varying complexities, as listed in Table \ref{tab:squad_zeroshot_prompts}. Our results, detailed in Table \ref{tab:squad_zeroshot_prompts}, reveal that our template achieves the highest F1 score. This indicates the lower performance of zero-shot Flan-T5 on SQuAD and similar MRC datasets is expected, even with the original instruct-tuning templates. It supports our hypothesis that, although Flan-T5 is instruct-tuned on SQuAD, its primary strengths are in broader generative question answering and reasoning, rather than specific extractive QA tasks such as MRC, particularly when evaluated by exact match and F1 metrics.

\begin{table}[!htbp]
    \centering
    \small
    \begin{tabular}{p{1in}|cc}
        \hline\hline
         & \multicolumn{2}{c}{\textbf{SQuAD Performance}} \\
        \textbf{Prompt Template} & \textbf{EM} & \textbf{F1} \\\hline\hline
        \begin{tabular}[c]{@{}l@{}}Article: \{context\}\\Question: \{question\}\\Answer:\end{tabular} & 7.001 & 21.717 \\\hline
        \begin{tabular}[c]{@{}l@{}}Answer a question\\about this article.\\Article: \{context\}\\Question: \{question\}\\Answer:\end{tabular} & 15.875 & 33.375 \\\hline
        \begin{tabular}[c]{@{}l@{}}Here is a question\\about this article:\\Article: \{context\}\\What is the answer\\to this question:\\Question: \{question\}\\Answer:\end{tabular} & \textbf{16.764} & 35.304 \\\hline\hline
        \begin{tabular}[c]{@{}l@{}}Our Template\\See Table \ref{tab:prompts}\end{tabular} & 16.149 & \textbf{37.691} \\\hline
    \end{tabular}
    \caption{Flan-T5-Large zero-shot performance on SQuAD with different prompt templates.}
    \label{tab:squad_zeroshot_prompts}
\end{table}

\subsection{Discussion on Llama 2 Performance}
\label{subsec:llama_performance}

We observe that models based on Llama 2 and Alpaca generally underperform compared to those based on Flan-T5, in both zero-shot and fine-tuned scenarios, with or without \textit{QASE}. This section delves into a detailed discussion of the potential reasons behind this trend.

Firstly, the discrepancy in performance may stem from the inherent structural differences between decoder-only models (Llama 2 and Alpaca) and encoder-decoder models (Flan-T5). Encoder-decoder models are better equipped for tasks that require extensive input processing, such as MRC, making them more apt for these tasks than decoder-only models, which are typically more suited to open-ended QA scenarios. This fundamental distinction partially accounts for Flan-T5's superior performance in context-based question answering across both zero-shot and fine-tuned settings.

Additionally, the difference in the number of trainable parameters during fine-tuning might contribute to the observed performance gap. Table \ref{tab:params} indicates that fine-tuning Llama 2 and Alpaca with LoRA leads to a significantly lower count of trainable parameters (4.2M) compared to even the smallest Flan-T5 model (77.0M). This disparity in trainable parameters is a crucial factor in explaining why fine-tuned Flan-T5 models, irrespective of the use of QASE, outperform Llama 2 and Alpaca models.

While we address these factors, conducting a comprehensive comparison and analysis of different generative model architectures in MRC tasks exceeds the scope of our current study. Nonetheless, we acknowledge that additional factors, such as the specific instruct-fine-tuning of Flan-T5 models on MRC datasets like SQuAD, might also play a role in their enhanced performance over Llama 2 and Alpaca.

\subsection{Discussion on Performance Discrepancy across Different Base PLMs and Datasets}
\label{subsec:performance_discrepancy}

\begin{table}[h]
    \centering
    \small
    \resizebox{0.48\textwidth}{!}
    {
        \begin{tabular}{c||c|c|c|c|c}
        \hline\hline
        & \textbf{Llama2} & \textbf{Alpaca} & \makecell[c]{\textbf{Flan-T5}\\ \textbf{Small}} & \makecell[c]{\textbf{Flan-T5}\\ \textbf{Base}} & \makecell[c]{\textbf{Flan-T5}\\ \textbf{Large}} \\
        \midrule
        & \multicolumn{5}{c}{$\Delta$EM} \\\\
        \textbf{SQuAD}     & 1.47  & 33.82  & 0.43          & 0.13          & 1.17          \\ 
        \textbf{MultiSpanQA}   & 1.61          & -1.01          & -0.08          & 0.32          & -0.73  \\
        \textbf{Quoref} & 19.24   & -    & 4.28      & 3.30      & 1.36      \\
        \midrule
        & \multicolumn{5}{c}{$\Delta$F1} \\\\
        \textbf{SQuAD}     & 1.34      & 8.35   & 0.46          & 0.76          & 1.09 \\ 
        \textbf{MultiSpanQA}   & 3.30    & 1.32          & 0.80          & 0.11   & 1.36  \\
        \textbf{Quoref} & 16.03   & -    & 5.66      & 0.35      & 2.04      \\\hline\hline 
        \end{tabular}
    }
    \caption{Performance improvement (in \%) of fine-tuned PLMs with \textit{QASE} on each dataset.}
    \label{tab:performance_discrepancy}
\end{table}

As shown in Table \ref{tab:performance_discrepancy}, we observe a significant performance improvement with \textit{QASE} across different base PLMs and datasets. Specifically, dataset-wise, a larger improvement is noted on Quoref compared to other datasets. This is partially due to the relatively weaker baseline performance on Quoref. For example, a fine-tuned Flan-T5-Large model without \textit{QASE} achieves an F1 score of 90.71\% on SQuAD, 83.09\% on MultiSpanQA, and 80.49\% on Quoref. Higher baseline scores indicate a strong initial performance, making further improvements more challenging and thus more meaningful. Despite the already high performance on the other two datasets, particularly SQuAD, the incorporation of \textit{QASE} still results in noticeable improvements.

PLM-wise, we generally observe that the improvements on Llama 2 and Alpaca are more substantial than those on the Flan-T5 base models, with few exceptions on MultiSpanQA. This trend can be partially attributed to the higher baseline performance of Flan-T5 models on these datasets. We discuss in Sections \ref{sec:discussions}, \ref{subsec:flan-t5_zeroshot}, and \ref{subsec:llama_performance} that factors such as (1) differences in pre-training datasets, with Flan-T5 models being fine-tuned on MRC tasks like SQuAD, and (2) varied adaptation to tasks due to structural disparities, can contribute to this performance gap. Encoder-decoder models, such as Flan-T5, are better equipped for tasks requiring extensive input processing, like MRC, making them more suitable for these tasks than decoder-only models, which are typically more suited to open-ended QA scenarios. This fundamental distinction partially accounts for Flan-T5's superior performance in context-based question answering across both zero-shot and fine-tuned settings. While acknowledging these factors, a comprehensive comparison of different generative model architectures in MRC tasks exceeds the scope of our study.

\end{document}